\documentclass[lettersize,journal]{IEEEtran}

\usepackage{amsmath,amssymb,amsfonts}
\usepackage{algorithm}
\usepackage{algpseudocode}
\usepackage{array}
\usepackage{textcomp}
\usepackage{stfloats}
\usepackage{url}
\usepackage{verbatim}
\usepackage{graphicx}
\usepackage{cite}
\usepackage{xcolor}
\usepackage{subfigure}
\usepackage{theorem}

\newtheorem{lemma}{Lemma}

\newtheorem{proposition}{Proposition}
\newtheorem{remark}{Remark}
\begin{document}
	
	\title{CARE: Confidence-rich Autonomous Robot Exploration using Bayesian Kernel Inference and Optimization}
	
	\author{Yang Xu, \IEEEmembership{Student Member, IEEE,} 
		Ronghao Zheng$^\dagger$, \IEEEmembership{Member, IEEE,}
		Senlin Zhang, \IEEEmembership{Member, IEEE,}\\
		Meiqin Liu, \IEEEmembership{Senior Member, IEEE,}
		and Shoudong Huang, \IEEEmembership{Senior Member, IEEE}
		\thanks{$\dagger$: Corresponding author.}
		\thanks{Yang Xu, Ronghao Zheng, and Senlin Zhang are with the College of Electrical Engineering, Zhejiang University, Hangzhou 310027, China, and also with the State Key Laboratory of Industrial Control Technology, Zhejiang University, Hangzhou 310027, China. (e-mail: {\{xuyang94, rzheng, slzhang\}@zju.edu.cn})
		}
		\thanks{Meiqin Liu is with the Institute of Artificial Intelligence and Robotics, Xi'an Jiaotong University, Xi'an 710049, China, and also with the State Key Laboratory of Industrial Control Technology, Zhejiang University, Hangzhou 310027, China. (e-mail: {liumeiqin@zju.edu.cn})}
		\thanks{Shoudong Huang is with the Robotics Institute, Faculty of Engineering
			and Information Technology, University of Technology Sydney, Ultimo, NSW 2007,
			Australia. (e-mail: shoudong.huang@uts.edu.au)}
	}
	
	\markboth{Journal of \LaTeX\ Class Files,~Vol.~14, No.~8, August~2021}%
	{Xu \MakeLowercase{\textit{et al.}}: CARE: Confidence-rich autonomous robot exploration using Bayesian kernel inference and optimization}
	
	
	\maketitle
	
	\begin{abstract}
		In this paper, we consider improving the efficiency of information-based autonomous robot exploration in unknown and complex environments. 
		We first utilize Gaussian process (GP) regression to learn a surrogate model to infer the confidence-rich mutual information (CRMI) of querying control actions, then adopt an objective function consisting of predicted {CRMI values and prediction uncertainties to conduct Bayesian optimization (BO), i.e., GP-based BO (GPBO). The trade-off between the best action with the highest CRMI value (exploitation)} and the action with high prediction variance (exploration) can be realized.
		To further improve the efficiency of GPBO, we propose a novel lightweight information gain inference method based on Bayesian kernel inference and optimization (BKIO), achieving an approximate logarithmic complexity without the need for training.
		BKIO can also infer the CRMI and generate the best action using BO with bounded cumulative regret, which ensures its comparable accuracy to GPBO with much higher efficiency.
		Extensive numerical and real-world experiments show the desired efficiency of our proposed methods without losing exploration performance in different unstructured, cluttered environments.
		We also provide our open-source implementation code at \textit{https://github.com/Shepherd-Gregory/BKIO-Exploration}.
	\end{abstract}
	
	
	\section{Introduction}
	
	\IEEEPARstart{R}{obot} exploration gains its prevalence recently in \textit{priori} unknown environments such as subterranean, marine, and planetary tasks
	\cite{Azpurua2021three,stankiewicz2021adaptive}.
	Among the literature, state-of-the-art exploration methods prefer to use information-theoretic metrics in each iteration, such as Shannon mutual information (MI) \cite{julian2014mutual} and its derivatives \cite{charrow2015information,zhang2020fsmi,yang2021crmi,xu2022confidence}, to evaluate the information gain brought by candidate control actions accurately. 
	For instance, in \cite{julian2014mutual}, Julian \textit{et al.} defined the Shannon MI between new observation and the occupancy grid map (OGM) at candidate poses. Charrow \textit{et al.} \cite{charrow2015information} adopted the Cauchy-Schwarz quadratic MI {(CSQMI)} to improve the time efficiency of Shannon MI in 2D and 3D exploration for computationally limited platforms. {These OGM-based MI metrics are also called occupancy grid MI (OGMI).} Our previous work \cite{yang2021crmi} further provided a more accurate information measure, confidence-rich mutual information (CRMI), over the dense confidence-rich map (CRM) \cite{agha2019confidence} for robot exploration. 
	
\begin{figure}
		\includegraphics[width=1.0\linewidth]{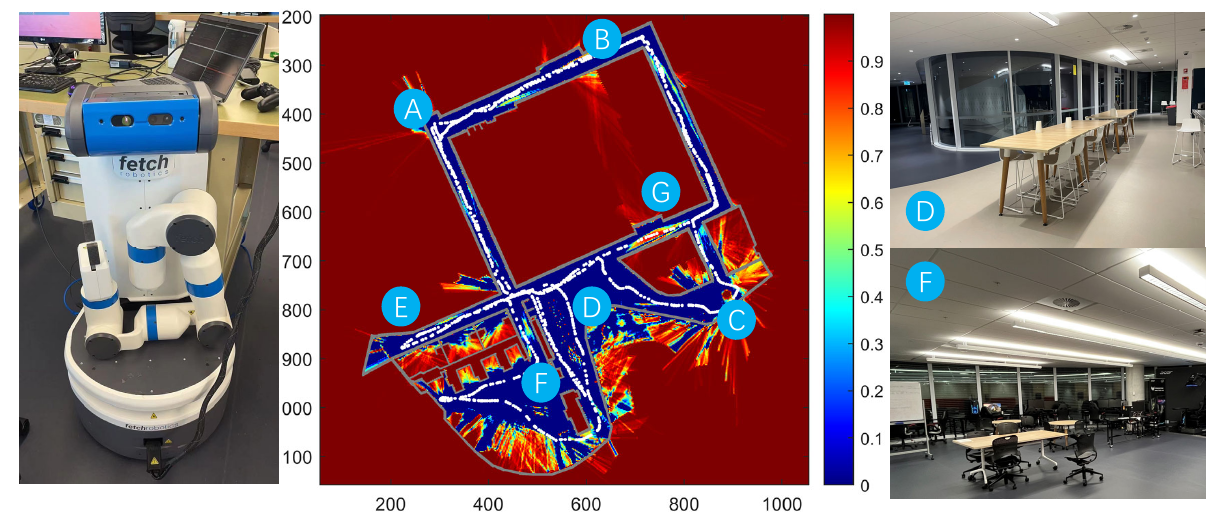}
		\centering
		\caption{CRMI-based autonomous robot exploration using a range-sensing Fetch robot with limited FOV in a cluttered office environment. \textbf{Left}: Fetch robot. \textbf{Middle}: Environmental CRMI surface after exploration. It shows the resulting exploration trajectory {(white)} in one experiment, where the robot's overall exploration sequence to maximize the CRMI is A-B-G-C-D-E-F-D-A-B-G-A. \textbf{Right}: Corresponding real-world views at points D and F. The color bar is the MI scale [0, 1] bit. Note that this informative trajectory is planned \textit{online} by the MI-driven planner. We set a high information threshold intentionally to explore rapidly. If setting a lower threshold, the CRMI-driven robot will explore the unexplored spaces (\textit{dark red}), then obscured spaces (\textit{light red}), and be repulsive to walls/obstacles (\textit{deep blue}). {The gray lines represent glass rooms, walls, or building boundaries the robot can not enter.}}
		\label{fig:fetch}
	\end{figure}

	Driven by these information metrics, the robot will choose and execute the most informative action, thus the exploration problem becomes a sequential optimal decision-making one naturally. According to \cite{julian2014mutual}, the good properties of MI also can ensure the accuracy and the eventual completeness of the exploration. 
	A real-world exploration example is in Fig.~\ref{fig:fetch}.
	
	Intuitively, the way to tackle an autonomous exploration problem is to use a greedy strategy and add more candidate actions, including sampled nodes \cite{hollinger2014sampling,ghaffari2019sampling}, available viewpoints \cite{ charrow2014approximate}, or special motion primitives \cite{charrow2015rss}, in the discrete action space. 
	However, the computational cost of the MI evaluation of all candidate actions will become expensive since the forward simulation in the evaluation requires extensive raycasting, map updates, and MI calculation. Notably, these consequences will be more distinct in 3D environments because the increased dimension needs to evaluate much more samples.
	
	\subsection{Related Work}
	\label{subsec:rel}
	In the context of robot exploration, deep learning has been introduced to realize predicting optimal sensing actions more efficiently. Bai \textit{et al.} \cite{bai2017toward} trained the deep neural network with plenty of randomly generated 2D gray maps to generate suggested action and ensure inferring in constant time. Graph neural networks have also been combined with reinforcement learning methods to learn the best action from an exploration graph, rather than metric maps or visual images \cite{chen2020autonomous,wang2018nervenet}. 
	Nevertheless, the neural network-based robot exploration methods require numerous offline training samples beforehand and are limited to the adaptability and generalization capability in different environments.
	
	More recently, statistical learning techniques also provide a powerful tool to find the global optimum approximately by training predictive models using minor parts of actions in continuous action spaces, without evaluating the objective function expensively, which also has better interpretability in black-box inference\cite{marchant2014bayesian,oliveira2020bayesian,francis2019occupancy}. 
	In \cite{bai2015inference}, Bai \textit{et al.} used the Gaussian process (GP) to model the relationship between control actions and the explicitly evaluated OGMI for the robot exploring \textit{priori} unknown areas.
	In \cite{bai2016information}, they further introduced Bayesian optimization (BO) into the OGMI-based robot exploration to optimize the GP prediction in multiple iterations, which provides rapid map entropy reduction and ensures computational efficiency. 
	Generally, BO assesses the acquisition function derived from the GP prior and samples, then chooses the next query point maximizing the acquisition function and balancing the trade-off between exploration (global) and exploitation (local). Iteratively, BO presents more precise results on the posterior distribution as the observations (training samples) increase.
	Rather than evaluating discrete viewpoints, Francis \textit{et al.}\cite{francis2019occupancy} modeled the autonomous exploration and mapping task as a constrained BO aiming to find optimal continuous paths.
	
	However, the main bottleneck of the GP-based methods is that the number of training actions $N$ will affect the resulting prediction accuracy directly, as well as the computational cost. That implies one needs to pay expensive computations to achieve higher exploration performance. Typically, updating and querying the GP models (the engine behind BO) have an overall $\mathcal{O}(N^3)$ time complexity. This compromises the inference efficiency and real-time performance of robot exploration tasks inevitably, especially in larger and 3D scenes.
	
	Notably, the Bayesian kernel inference (BKI) technique proposed in \cite{vega2014nonparametric} gives us a chance to perform efficient exact inference on a simplified model, rather than approximating inference on an exact generative model (e.g. GP) expensively. BKI extends local kernel estimation to Bayesian inference for exponential likelihood functions, enabling only $\mathcal{O}(\log N_q)$ ($N_q$: the number of querying samples) run time for inference. These significant merits enhance BKI's application in robotics, including sensor uncertainty estimation \cite{valentin2016predictive}, high-speed navigation \cite{richter2015bayesian}, as well as environment mapping using sparse sensor measurements such as terrain traversability mapping \cite{shan2018bayesian}, 3D occupancy mapping \cite{doherty2019learning}, semantic mapping\cite{lu2020bayesian}. 
	\subsection{Motivations and Contributions}
	
	{Motivated by the recent BKI technique \cite{vega2014nonparametric} and its robotic applications \cite{shan2018bayesian,doherty2019learning,gan2020bayesian}, in this paper, we aim to utilize it to accelerate the computationally expensive CRMI evaluation of control actions considerably without losing CRMI prediction accuracy and exploration performance in complex scenes and use BO to optimize the decision-making process.}
	Our main contributions are three-fold: 
	
	1) We first present a regular CRMI inference method based on GP and BO (GPBO) using minor explicitly evaluated samples to predict CRMI values and uncertainties of candidate actions, and use an information-theoretic objective function to realize the trade-off of the exploration-exploitation dilemma in MI prediction, rather than the naive-greedy (NG) method evaluating all candidate actions exhaustively and expensively;
	
	2) We further propose a novel Bayesian kernel inference and optimization method (BKIO) for CRMI inference without training a surrogate evaluation model. Compared with the GPBO method, BKIO can perform much more efficient MI values and uncertainties prediction in approximate logarithm time with comparable accuracy and exploration performance; 
	
	3) We conduct extensive numerical simulations in synthetic and dataset maps, and real-world experiments in a cluttered office scene to validate our proposed methods. We also release an open-source implementation of our proposed methods\footnote{https://github.com/Shepherd-Gregory/BKIO-Exploration}.
	
	The paper's organization is as follows. We formulate the GPBO-based CRMI inference problem in Section~\ref{sec:bo} and present BKIO for CRMI in Section~\ref{sec:BKIO}. Simulation results using synthetic data and the real-world experimental results, as well as the discussion, are given in Section~\ref{sec:simu}, followed by conclusions in Section~\ref{sec:con}.
	
	\section{BO-based CRMI-driven Robot Exploration}
	\label{sec:bo}
	
	In this paper, for simplicity of discussion, we mainly consider the CRMI-based information-theoretic exploration using a mobile robot equipped with a beam-based range sensor of a limited field of view (FOV) in 2D environments. 
	We aim to evaluate the exact CRMI of unknown robot configurations sampled in the action space efficiently.
	
	\subsection{Information-Theoretic Exploration using CRMI}
	Considering the information-based autonomous robot exploration in an unknown, cluttered, and unstructured environment, we prefer to use CRMI as the information measure, since CRMI is more descriptive and accurate in cluttered and unstructured areas than the OGMI derived from Bernoulli distribution-based OGMs, especially when using sparse measurements or low-resolution sensors \cite{yang2021crmi}. 
	This advantage stems from the underlying CRM, a dense environmental representation consisting of non-parametric occupancy distribution over map cells \cite{agha2019confidence}, which can encode the measurement dependencies between map cells within a sensor cone into the map belief over each cell, thereby enhancing the perception ability of the robot in uncertainty-aware navigation and planning tasks.
	
	Here we begin with some related definitions and notions.
	Generally, the robot generates a set of candidate actions  $\mathcal{X}_{\mathrm{action}}$ in the robot's feasible configuration space $\mathcal{X}\subseteq SE(2)$. For a candidate configuration $x_i=[p_i^x, p^y_i, \psi_i] \in \mathcal{X}_{\mathrm{action}}$, $p_i^x$ and $p^y_i$ denote the robot's position on the map, and $\psi_i$ denotes the heading angle of the robot. 
	We assume this configuration space has been discretized by a fixed resolution over the 2D static grid map comprised of $N_m$ grid cells. The continuous occupancy values over this map are $m=\{m^{[1]},\dots,m^{[\xi]},\dots,m^{[N_m]}\}(m^{[\xi]} \in [0,1])$,  i.e. the occupancy level over the independent grid cells, which can be updated and queried by the CRM method. 
	
	In a typical CRM, the map belief $b^{m^{[\xi]}}_t:=p({m^{[\xi]}}|z_{1:t},x_{1:t})$ of a cell $m^{[\xi]}$ at time $t$ is a non-parametric continuous occupancy distribution {conditioned upon} the sensor observations $z_{1:t}$ and robot poses $x_{1:t}$. The expected map occupancy over a cell $m^{[\xi]}$ can be defined as the mathematical expectation of the map belief $b^{m^{[\xi]}}$:
	$
	\hat{m}^{[\xi]}:=\mathbb{E}[{m^{[\xi]}}]=\int_0^1{{m^{[\xi]}}b^{m^{[\xi]}}}dm^{[\xi]},
	\label{eq:hatm}
	$
	and the confidence of map occupancy can be defined as the variance of map belief distribution.
	The occupancy distribution of an unobserved map cell $m^{[\xi]}$ is assumed to be uniform, i.e., $\hat{m}^{[\xi]} = 0.5$. This unobserved cell contributes an entropy of 1 bit and a totally known cell (i.e. $\hat{m}^{[\xi]} = $ 0 or 1) contains 0-bit entropy according to the information theory. 
	
	From the view of information theory, the expected information gain of $x_i$ can be evaluated by the current map entropy and conditional entropy given a new measurement at $x_i$:
	\begin{equation}
		I(m;x_i) = H(m) - H(m|x_i).
		\label{eq:mi}
	\end{equation}
	
	The aim of information-theoretic robot exploration is to select the best action $x_{\mathrm{best}}$ maximizing the expected CRMI:
	\begin{equation}
		x_{\mathrm{best}}=\mathop{\mathrm{argmax}}\limits_{x_i\in \mathcal{X}_{\mathrm{action}}} I(m;x_i).
		\label{eq:maxmi}
	\end{equation}
	
	Generally, the expected CRMI of a candidate configuration $x_i$ can be calculated by the \textit{virtual} measurement of a scan $z'$ raycasting on currently built map $m$ at pose $x_i$, then updating a virtual map $m'$ and decomposing the CRMI of a scan into each cell on each independent beam, thus the CRMI can be accumulated over all cells on the map approximately, so Eq.~\eqref{eq:mi} can be rewritten as follows:
	\begin{equation}
		\begin{split}
			I(m;x_i) = H(m) - H(m'|x_i,z') \\
			\approx \sum_{j=1}^{N_z}\sum_{k\in\mathcal Z_{[j]}} I(m'_{[k]};z'_{[j]}),
		\end{split}
		\label{eq:summi}
	\end{equation}
	where $N_z$ is the number of beams of the virtual scan $z'$, $\mathcal Z_{[j]}$ is the index set of map cells intersected with the $j$th beam $z'_{[j]}$ of scan $z'$.
	
	The CRMI of the cell $m'_{[k]}$ with the measurement beam $z'_{[j]}$ can be calculated by the following equation using numerical integration:
	\begin{align}\label{eq:crmi}
		&I(m'_{[k]};z'_{[j]})
		=H(m'_{[k]}|z_{1:t-1}) - H(m'_{[k]}|z'_{[j]},z_{1:t-1}) \nonumber \\ 
		=&\int_{z\in {Z_m}}p(z'_{[j]}|z_{1:t-1})\int_0^1 b_{t}^{m'_{[k]}}\log b_{t}^{m'_{[k]}} dm'_{[k]} dz'_{[j]}  \nonumber \\
		-&\int_0^1 b_{t-1}^{m'_{[k]}}\log b_{t-1}^{m'_{[k]}} dm'_{[k]},
	\end{align}
	where $Z_m$ is the maximum measurement range. 
	For more details about CRMI calculation, please see \cite{yang2021crmi}.
	
	\begin{proposition}\label{prop:mieval}
		{\cite{yang2021crmi}} The CRMI calculation process involves numerous numerical integration and accumulation operations. Hence, the overall computational complexity of evaluating a configuration is about $O(N_z N_c^2)$ for updating the map and $O(N_z N_c^2 \lambda_z^{-1}\lambda_m^{-1})$ for the MI calculation, where $\lambda_z$ and $\lambda_m$ are the numerical integration resolutions in Eq.~\eqref{eq:crmi}, $N_c$ is the maximum size of $\mathcal Z_{[j]}$.
	\end{proposition}
	
	The explicit CRMI evaluation of all sampled actions imposes notable computational costs for robot exploration, especially in unstructured, cluttered, and 3D scenes, since the robot needs to assess more actions to keep safety and exploration performance in these cases. 
	
	\subsection{BO-based CRMI Inference for Robot Exploration}
	\label{subsec:gpbo}
	To alleviate the computational burden of evaluating numerous actions, here we first introduce GP and BO to infer the CRMI with a training set of explicitly evaluated actions.
	
	Consider a supervised learning-based inference problem on predictive stochastic models $p(\mathbf{y}|\mathbf{x})$ given a sequence of $N$ explicitly evaluated samples $\mathcal{D}=\{(\mathbf{x}=\{x_i\},\mathbf{y}=\{y_i\})\}_{i=1}^N$, where $\mathbf{x}$ and $\mathbf{y}$ represent the set of evaluated configurations and the resulting CRMI values $I(m;\mathbf{x})$, respectively. 
	The main objective is to infer the posterior distribution $p(\mathbf{y}^*|\mathbf{x}^*,\mathcal{D})$ to evaluate CRMI values $\mathbf{y}^*\in \mathbb{R}^{N_q}$ of the $N_q$ querying sample inputs $\mathbf{x}^*\in \mathcal{X}_{\mathrm{action}}$.
	
	Here we assume $p(\mathbf{y}|\mathbf{x})$ follows a GP, according to \cite{rasmussen2005gaussian}, we can estimate the posterior mean $\bar{\mathbf y}^*$ and covariance $cov(\mathbf y^*)$ of the output of the querying set of candidate actions $\mathbf{x}^*$, i.e.:
	\begin{align}\label{eq:gpmivar}
		\bar{\mathbf y}^* &= k(\mathbf{x}^*,\mathbf{x})[k(\mathbf{x},\mathbf{x})+\boldsymbol{\sigma}_N^2 \mathbf{I}]^{-1}\mathbf{y}, \\
		cov(\mathbf y^*) &= k(\mathbf{x}^*,\mathbf{x}^*)-k(\mathbf{x}^*,\mathbf{x})[k(\mathbf{x},\mathbf{x})+\boldsymbol{\sigma}_N^2 \mathbf{I}]^{-1}k(\mathbf{x},\mathbf{x}^*), \nonumber
	\end{align}
	where $\boldsymbol{\sigma}_N^2$ is the Gaussian noise variance vector associated with input $\mathbf{x}$, $k(\cdot,\cdot)$ is a kernel function measuring the proximity between two elements in feature space. 
	
	{Among the available kernel functions, we prefer Mat$\acute{\text e}$rn kernel for its capability of handling sudden transitions of terrain \cite{rasmussen2005gaussian,ramos2012gaussian} since the potential obstacles and unknown structures in application scenes that have never been seen before will vary the MI values greatly.}

	In practice, we choose a Mat$\acute{\text e}$rn 5/2 kernel with the form as:
	\begin{equation}
		k(\mathbf{x}^*,\mathbf{x}) = (1+\frac{\sqrt{5}r}{\ell}+\frac{5r^2}{3\ell^2})\exp (-\frac{\sqrt{5}r}{\ell}), r=||\mathbf{x}^*-\mathbf{x}||,
		\label{eq:matern}
	\end{equation}
	where $\ell$ is a positive characteristic length scale.

	At each exploration step, the information-driven robot expects to choose the actions with high predicted CRMI values (\textit{exploitation}) to maximize the information gain locally (i.e. Eq.~\eqref{eq:maxmi}). Still, the results directly predicted by GP tend to recommend sub-optimal action because of the insufficient evaluated samples and the resulting inaccurate predictive model. Hence, the actions with high predicted uncertainty can also not be neglected in the inference process (\textit{exploration}).
	
	To handle this, we incorporate the prediction uncertainty of CRMI values to the objective function in Eq.~\eqref{eq:maxmi} to realize a local trade-off between exploration and exploitation. {Referring to the GP-UCB in \cite{srinivas2010gaussian}, here} we can get the suggested action maximizing the information objective function $f^I$ based on Eq.~\eqref{eq:maxmi} and Eq.~\eqref{eq:gpmivar} after multiple epochs of optimization:
	\begin{equation}
		x_{s}=\mathop{\mathrm{argmax}}\limits_{\mathbf{x}\in \mathcal{X}_{\mathrm{action}}} f^I,
		f^I= \bar{\mathbf y}+ \alpha^{1/2}\sigma_{\mathbf y},
		\label{eq:ucb}
	\end{equation}
	where $\alpha$ is the trade-off factor, $\sigma_{\mathbf y}$ is the standard deviation of CRMI prediction.
	
	For such a sequential multi-armed bandit problem \cite{auer2002finite}, one can use the expected regret to ensure that the multi-epoch optimization is reasonable and practical, i.e. the regret of the upper confidence bound (UCB) function should be bounded over rounds. 
	\begin{lemma} \cite{srinivas2010gaussian}
		For finite $|\mathcal{D}|<\infty$, let $\delta\in(0,1)$ and $\alpha=2\log(|\mathcal{D}|T^2\pi^2/6\delta)$, the UCB with $\alpha$ for a sample of a GP with mean function zero and covariance function $k(\mathbf{x}^*,\mathbf{x})$, can obtain a regret bound of $\mathcal{O}^*(\sqrt{N_{\mathrm{epoch}}\gamma_T\log |\mathcal{D}|})$ with high probability, where $N_{epoch}$ is the optimization rounds, $T\in[1, N_{\mathrm{epoch}}]$ is the current optimization round, and $\gamma_T$ is the max information gain after $N_{\mathrm{epoch}}$ rounds.
	\end{lemma}
	\begin{remark}
		According to Lemma 1, the GP-based BO using the UCB of Eq.~\eqref{eq:ucb} and the Mat$\acute{\text e}$rn 5/2 kernel can achieve the desired trade-off of exploration and exploitation after $N_{\mathrm{epoch}}$ rounds optimization. 
	\end{remark}
	
	\section{Bayesian Kernel Inference and Optimization for CRMI-based Robot Exploration}
	\label{sec:BKIO}
	In this section, we present a novel and highly efficient CRMI inference method based on the BKI and BO for autonomous robot exploration.
	
	\subsection{Bayesian Generalized Kernel Inference}
	\label{subsec:bki}
	Here we continue to use the dataset $\mathcal{D}$ and the querying set $\mathbf{x}^*$ in Section \ref{subsec:gpbo}, and assume the CRMI output $\mathbf{y}^*$ and the actions input $\mathbf{x}^*$ follows the relationship of $p(\mathbf{y}|\mathbf{x})$ (not a GP model). {According to {the BKI theory and applications \cite{vega2014nonparametric,shan2018bayesian,gan2023multitask}}, the Bayesian generalized kernel inference problem can be formulated into three steps:}
	
	{1) \textit{\textbf{Target parameters inference}}: First, this problem can be solved by associating latent parameters $\boldsymbol \theta=\{\theta_i\}_{i=1}^N \in \mathbf \Theta$ with input $\mathbf{x}$ in the latent space $\mathbf \Theta$, where the likelihood $p(\mathbf{y}|\boldsymbol{\theta})$ is known. Thus the inference on $\mathbf{y}^*$ can be formulated as an inference on target parameters $\boldsymbol{\theta}^* \in \mathbf \Theta$ related to $\mathbf{x}^*$:
		\begin{equation}
			p(\mathbf{y}^*|\mathbf{x}^*,\mathcal{D}) = \int_{\mathbf \Theta} p(\mathbf{y}^*|\boldsymbol{\theta}^*) p(\boldsymbol{\theta}^*|\mathbf{x}^*,\mathcal{D}) d\boldsymbol{\theta}^*,
			\label{eq:post}
		\end{equation}
		where the posterior distribution of the latent variables can be characterized using Bayes' rule:
		$
		p(\boldsymbol{\theta}^*|\mathbf{x}^*,\mathcal{D}) \propto \int_{\mathbf \Theta} \prod_{i=1}^{N} p(y_i|{\theta}_i)p(\boldsymbol{\theta}_{1:N},\boldsymbol{\theta}^*|\mathbf{x}_{1:N},\mathbf{x}^*)d {\boldsymbol{\theta}_{1:N}}.
		$ }
	
	{2) \textit{\textbf{Latent variables marginalization}}: By strongly assuming latent parameters $\boldsymbol{\theta}_{1:N}$ are conditionally independent given the target parameters $\boldsymbol{\theta}^*$:
		$	p(\boldsymbol{\theta}_{1:N},\boldsymbol{\theta}^*|\mathbf{x}_{1:N},\mathbf{x}^*) = \prod_{i=1}^{N} p({\theta}_i|\boldsymbol{\theta}^*,{x}_i,\mathbf{x}^*) p(\boldsymbol{\theta}^*|\mathbf{x}^*),
		$
		one can marginalize the latent variables $\boldsymbol{\theta}_{1:N}$ and then obtain 
		$
		p(\boldsymbol{\theta}^*|\mathbf{x}^*,\mathcal{D}) \propto \prod_{i=1}^{N} p(y_i|\boldsymbol{\theta}^*,{x}_i,\mathbf{x}^*) p(\boldsymbol{\theta}^*|\mathbf{x}^*).
		$ }
	
	{3) \textit{\textbf{Kernel approximation}}: BKI further defines a distribution that has a special smoothness constraint and bounded Kullback-Leibler divergence $D_{KL}(g||f)$ between the extended likelihood $p(y_i|\boldsymbol{\theta}^*,{x}_i, \mathbf{x}^*)$ represented by $g$ and the likelihood $p(y_i|{\theta}_i)$ represented by $f$, i.e., the maximum entropy distribution $g$ satisfying $D_{KL}(g||f)\leq \rho(\mathbf{x}^*,\mathbf{x})$ has the form $g(\mathbf{y})\propto f(\mathbf{y})^{k(\mathbf{x}^*,\mathbf{x})}$, where $\rho(\cdot,\cdot):\mathcal{X}\times \mathcal{X}\to \mathbb{R}^+$ is a smoothness bound and $k(\cdot,\cdot):\mathcal{X}\times \mathcal{X}\to [0,1]$ is the kernel function which can be uniquely determined by $\rho$. 
		Substituting it into Eq.~\eqref{eq:post} yields:
		\begin{equation}
			p(\boldsymbol{\theta}^*|\mathbf{x}^*,\mathcal{D}) \propto \prod_{i=1}^{N} p(y_i|\boldsymbol{\theta}^*)^{k(\mathbf{x}^*,\mathbf{x})} p(\boldsymbol{\theta}^*|\mathbf{x}^*).
			\label{eq:exp}
		\end{equation}
	}
	
	Thus, the posterior distribution can be exactly inferred by using the likelihood from the exponential family (e.g. Gaussian distribution) and assuming the corresponding conjugate prior.
	
	\subsection{BKI Optimization for Autonomous Robot Exploration}
	\label{subsec:bkio}
	{Here we present the BKIO method for information-driven robot exploration, using BO to optimize BKI results in multiple epochs.}
	
	{1) \textit{\textbf{Bayesian kernel CRMI inference}}: According to Section~\ref{subsec:bki}, we assume the underlying likelihood $p(\mathbf{y}|\boldsymbol{\theta})$ follows a Gaussian distribution with an unknown mean vector $\boldsymbol{\theta}$ and a fixed, known covariance matrix function  $\boldsymbol{\Sigma}(\mathbf{x})$:}
	\begin{equation}
		\mathbf{y} \sim \mathcal{N} (\boldsymbol{\theta}, \boldsymbol{\Sigma}(\mathbf{x})), \boldsymbol{\Sigma}(\mathbf{x})=\mathrm{diag}(\sigma_{\theta}^2)\in \mathbb{R}^{N\times N},
		\label{eq:muy}
	\end{equation}
	where $\sigma_{\theta}^2$ is the known variance associated with $\boldsymbol{\theta}$. Thus the conjugate prior of $p(\mathbf{y}|\boldsymbol{\theta})$ can also be described by a Gaussian distribution using the hyperparameter $\zeta$ and samples input $\mathbf{x}$:
	\begin{equation}
		p(\boldsymbol{\theta}|\mathbf{x}) = \mathcal{N} \left(\boldsymbol{\theta}_0, {\boldsymbol{\Sigma}(\mathbf{x})}/{\zeta}\right),
		\label{eq:mux}
	\end{equation}
	where $\boldsymbol{\theta}_0$ and $\zeta$ are the initial belief of the mean and the uncertainty of that belief, respectively. $\boldsymbol{\theta}_0$ is a vector consisting of $N$ known equal constants $\theta_0$, and $\zeta=0$ means no confidence and $\zeta\to \infty$ indicates full prior knowledge. 
	Here we assume $\zeta$ is a quite small positive constant since we do not have much prior information about the belief when exploring unknown areas.  
	Therefore, since the querying samples set $\mathcal{D}^*=\{(\mathbf{x}^*=\{x_i^*\},\mathbf{y}^*=\{y_i^*\})\}_{i=1}^{N_q}$ also follows the distributions of Eq.~\eqref{eq:muy} and Eq.~\eqref{eq:mux}, given the set of explicitly evaluated samples $\mathcal{D}$, 
	substituting Eq.~\eqref{eq:mux} and Eq.~\eqref{eq:muy} into Eq.~\eqref{eq:exp} yields:
	\begin{eqnarray}
		p(\boldsymbol{\theta}^*|\mathbf{x}^*,\mathcal{D}) \propto \prod_{i=1}^{N} \exp \left(- \frac{(y_i-\theta_i)^2}{2\sigma_{\theta}^2} k(\mathbf{x}^*,\mathbf{1}_{N_q}\times{x}_i)\right) \\ \nonumber
		\cdot \exp \left(-\frac{1}{2} \frac{(\theta_i-\theta_0)^2}{\sigma_{\theta}^2} \zeta \right),
	\end{eqnarray}
	thus the predicted mean and covariance of the CRMI posterior can be derived as follows:
	\begin{align}
		&\bar{\mathbf y}^*=\mathbb{E}[\mathbf y^*|\mathbf{x}^*,\mathcal{D}]=\mathbb{E}[\boldsymbol{\theta}^*|\mathbf{x}^*,\mathcal{D}]=\frac{\overline{\mathbf{y}}+\zeta\theta_0}{\zeta+\overline k}\simeq \frac{\overline{\mathbf{y}}}{\overline{k}}, \nonumber \\ 
		&cov({\mathbf y}^*)=\mathbb{V}[\boldsymbol{\theta}^*|\mathbf{x}^*,\mathcal{D}]=\frac{\boldsymbol{\Sigma}(\mathbf{x}^*)}{\zeta+\overline k} \simeq \frac{\boldsymbol{\Sigma}(\mathbf{x}^*)}{\overline k}, 
		\label{eq:mivar}
	\end{align}
	where $\overline{\mathbf{y}}$ and $\overline{k}$ can be computed by the kernel function:
	\begin{equation}
		\overline{k} = \Sigma_{i=1}^N k(\mathbf{x}^*,\mathbf{x}),~
		\overline{\mathbf{y}} = \Sigma_{i=1}^N k(\mathbf{x}^*,\mathbf{x}) y_i.
		\label{eq:barky}
	\end{equation}
	
	Therefore, given a set $\mathcal{D}$ of explicitly evaluated samples as the input dataset, we can easily compute the CRMI and the corresponding prediction uncertainty for the querying configurations $\mathbf{x}^*$ by using Eq.~\eqref{eq:mivar} and Eq.~\eqref{eq:barky}. 
	
	{2) \textit{\textbf{BKI optimization}}: To optimize the BKI inference results of CRMI, we apply Bayesian optimization and use the objective function Eq.~\eqref{eq:ucb} to ensure the trade-off between exploration and exploitation at each step. 
	}
	
	\begin{remark}
		According to \cite{srinivas2010gaussian}, as a special finite case of GP, the multivariate Gaussian posterior over $\mathbf{y}^*$ inferred by BKIO (c.f. Eq.~\eqref{eq:mivar}) can also ensure the Bayesian optimization regret is bounded using the Mat$\acute{\text e}$rn 5/2 kernel function and the UCB function of $f^I$ in Eq.~\eqref{eq:ucb}, which improves the optimality of final suggested action after multiple epochs optimization in each exploration step.
	\end{remark}
	
	The autonomous exploration framework based on the BKIO method is given in Algorithm \ref{alg:bkiexp}, where the function `GenActions' (Line 5) generates $N+N_q$ actions uniformly in the robot's current sensing range, `ComputeCRMI' (Line 12) refers to Eq.~\eqref{eq:summi} and Eq.~\eqref{eq:crmi}, `BKIOptimization' (Line 16) is in the Algorithm \ref{alg:bkiopt}. In each iteration, the robot first generates feasible actions $x_{\mathrm{cand}}$ in its sensing range, e.g., using frontier-based or sampling-based methods, then computes the CRMI values of the actions $\mathbf{x}$ randomly sampled from $x_{\mathrm{cand}}$ (Lines 5-14). The robot uses Algorithm~\ref{alg:bkiopt} to obtain the sets of best actions and the associated MI values and chooses the action with the maximum MI if it's higher than the threshold, otherwise, the robot will go back to the last action to end the expansion in this branch (Lines 16-23).
	Once the best action is confirmed, the robot will use the classic A* algorithm to plan the local collision-free path from the current pose to the best action, and then update the global map using the measurements and poses along the planned local path (Lines 25-26). The exploration will end when the iteration counts meet $N_{\mathrm{loop}}$ or the action history set is empty.
	
	\begin{algorithm}[ht] 
		\caption{BKIO-Exploration(~)} 
		\begin{algorithmic}[1]
			\Require {Occupancy map at $k$th time step $m_k$, previous robot poses $x_{\mathrm{hist}}=x_{0:k-1}$ and current pose $x_k$, the number of explicitly evaluated samples $N$, information threshold $I_{\mathrm{th}}$, the number of querying samples $N_{q}$, loop limit $N_{\mathrm{loop}}$}
			\State {iter = 0}
			\While {$x_{\mathrm{hist}} \neq \emptyset$ OR iter $<N_{\mathrm{loop}}$}
			\State {iter = iter + 1;}
			\State //~\textit{Generate $N+N_q$ actions uniformly near $x_k$}
			\State {$\mathbf x_{\mathrm{cand}} \leftarrow$\textbf{GenActions}$(x_k, m_k)$;}
			\State {//~\textit{Sample $N$ actions randomly from $\mathbf x_{\mathrm{cand}}$}}
			\State {$\mathbf x \leftarrow$\textbf{RandSample}$(\mathbf x_{\mathrm{cand}},N)$;}
			\State {$\mathbf x^* \leftarrow \mathbf x_{\mathrm{cand}}\setminus{\mathbf x};$ // \textit{The set of $N_q$ querying actions}}
			\State //~\textit{Evaluate these reference actions explicitly}
			\For {each $x_i\in \mathbf x$}
			\State {$m_{\mathrm{vir}}\leftarrow$\textbf{Raycasting}$(x_i,m_k)$;}
			\State {$I_i \leftarrow$\textbf{ComputeCRMI}$(m_{\mathrm{vir}},x_i)$;}
			\State {$\mathbf y \leftarrow \mathbf y \cup I_i$;}
			\EndFor
			\State //~\textit{Find the suggested action using Algorithm~\ref{alg:bkiopt}}
			\State {$\{x_{\mathrm{best}}, I_{\mathrm{best}}\} \leftarrow$\textbf{BKIOptimization}$(\{\mathbf x,\mathbf y\},\mathbf x^*,m_k)$;}
			\If{\textbf{Max}$(I_{\mathrm{best}})>I_{\mathrm{th}}$}
			\State {$x_{k+1} \leftarrow x_{\mathrm{best}}$(MaxInfoIndex);}
			\State {$x_{\mathrm{hist}} \leftarrow x_{\mathrm{hist}} \cup x_{k+1}$;}
			\Else 
			\State {$x_{k+1} \leftarrow x_{k-1};$ // \textit{Back to previous action}}
			\State Remove $x_{k-1}$ from $x_{\mathrm{hist}}$;
			\EndIf
			\State // \textit{Execute the action and update the map}
			\State $\mathcal{P}_{\mathrm{local}}\leftarrow$\textbf{Astar}$(x_k,x_{k+1})$ // \textit{Plan local path by A*}
			\State {$m_{k+1} \leftarrow $\textbf{ConfidenceRichMapping}$(\mathcal{P}_{\mathrm{local}})$;}
			\EndWhile 
		\end{algorithmic} 
		\label{alg:bkiexp}
	\end{algorithm}
	In Algorithm~\ref{alg:bkiopt}, for each optimization epoch, we first compute the kernel function matrix and use it to compute the CRMI values with the covariance of the querying set (Lines 4-7). Then we can obtain the objective function values of all querying actions and the action with the highest objective function value (Lines 8-9). We can add it into the best action set if it is already in $\mathcal{D}$; otherwise, we will evaluate it explicitly using Eq.~\eqref{eq:summi} and add it into $\mathcal{D}$ to enlarge the reference dataset and enhance the prediction accuracy (Lines 10-18).
	
	\begin{algorithm}[ht] 
		\caption{BKIOptimization(~)} 
		\begin{algorithmic}[1]
			\Require {Occupancy map at $k$th time step $m_k$, the set of explicitly evaluated samples $\mathcal{D}=\{(x_i, y_i)\}_{i=1}^N$, the set of querying actions $\mathbf x^*$, the number of optimization epoch $N_{\mathrm{epoch}}$, factor $\alpha$}
			\State {$x_{\mathrm{best}} \leftarrow \{\}, f^I_{\mathrm{best}} \leftarrow \{\}$;}
			\For {each epoch}
			\State //~\textit{Compute the kernel function using} Eq.~\eqref{eq:matern}
			\State {$k \leftarrow$\textbf{KernelFunction}$(\mathbf x^*,\mathbf x)$;}
			\State //~\textit{Infer CRMI and uncertainty using} Eq.~\eqref{eq:mivar}
			\State $\overline{k} \leftarrow \Sigma k,~\overline{y} \leftarrow k\cdot\mathbf y$;
			\State $\mathbf{y}^* \leftarrow \overline{y}/\overline{k},~cov(\mathbf{y}^*) \leftarrow \Sigma/\overline{k}$;
			\State $f^I \leftarrow \mathbf{y}^*+ \alpha^{1/2}\sigma_{\mathbf{y}^*}$;
			\State {$[x_s, \text{idx}] \leftarrow $\textbf{Max}$(\{f^I,\mathbf{y}\})$;}
			\If{$x_s \notin \mathbf x$}
			\State // \textit{Evaluate the CRMI of $x_s$ explicitly}
			\State {$m_{\mathrm{vir}}\leftarrow$\textbf{Raycasting}$(x_s,m_k)$;}
			\State $y_s =~$\textbf{ComputeCRMI}$(m_{\mathrm{vir}},x_s)$;
			\State // \textit{Add into $\mathcal{D}$ to improve prediction accuracy}
			\State $\mathbf x \leftarrow \mathbf x \cup x_s, \mathbf y \leftarrow \mathbf y \cup y_s$, {$\mathbf x^* \leftarrow \mathbf x^* \setminus{x_s};$} 
			\Else 
			\State {$y_s \leftarrow \mathbf{y}$(idx);}
			\EndIf
			\State {// \textit{Save actions with highest objective function values}}
			\State {$x_{\mathrm{best}} \leftarrow x_{\mathrm{best}} \cup x_s, I_{\mathrm{best}} \leftarrow I_{\mathrm{best}} \cup y_s$}
			\EndFor 
			\\
			\Return $x_{\mathrm{best}}, I_{\mathrm{best}}$
		\end{algorithmic} 
		\label{alg:bkiopt}
	\end{algorithm}
	
	\begin{proposition} \label{prop:bkio}
		The time complexity of our proposed BKIO method for evaluating all generated actions in Algorithm \ref{alg:bkiexp} and \ref{alg:bkiopt} is $\mathcal{O}(N(N_zN_c^2+ N_z N_c^2 \lambda_z^{-1}\lambda_m^{-1}))$ of explicit CRMI evaluation and $\mathcal{O} (N_{\mathrm{epoch}}N\log{N_{q}})$ of BKIO. The GPBO-based CRMI exploration methods in Section~\ref{subsec:gpbo} have the same time complexity of explicit CRMI evaluation as BKIO but a complexity of $\mathcal{O}(N_{\mathrm{epoch}}(N^3+N^2 N_{q}))$ to perform the expensive GP inference for CRMI,
		where $N_{epoch}$ is the number of optimization epochs.
	\end{proposition}
	
	Comparatively, the NG-based exploration method which evaluates all actions explicitly has a time complexity of $\mathcal{O}((N+N_q)(N_zN_c^2+ N_z N_c^2 \lambda_z^{-1}\lambda_m^{-1}))$ for explicit CRMI evaluation according to \textbf{Proposition} \ref{prop:mieval}. The above comparative results indicate that the BKIO-based exploration method significantly outperforms GPBO and NG methods in time efficiency, especially in large-scale and cluttered places that need more samples to evaluate rapidly (much greater $N$ and $N_{q}$).
	
	\section{Results and Discussions}
	\label{sec:simu}
	In this section, we aim to validate the performance of the proposed BKIO method,
	compared with other famous exploration methods such as naive greedy-based (``NG'') \cite{yang2021crmi}, and GPBO with multiple epochs (``GPBO''), where GPBO replaces Line 16 of Algorithm \ref{alg:bkiexp} with the function `GPBOptimization' using Eq. \eqref{eq:gpmivar}. NG evaluates all actions ($\mathbf x$ and $\mathbf x^*$) explicitly using Algorithm \ref{alg:bkiexp} (except Lines 15-16). Note that \textbf{NG} in \cite{yang2021crmi} can be treated as the \textit{ground truth} for verifying the CRMI prediction accuracy and exploration performance. 
	
	\begin{figure}
		\centering
		\subfigure[Structured synthetic map.]{
			\begin{minipage}[t]{0.5\linewidth}
				\centering
				\includegraphics[width=1.0\linewidth]{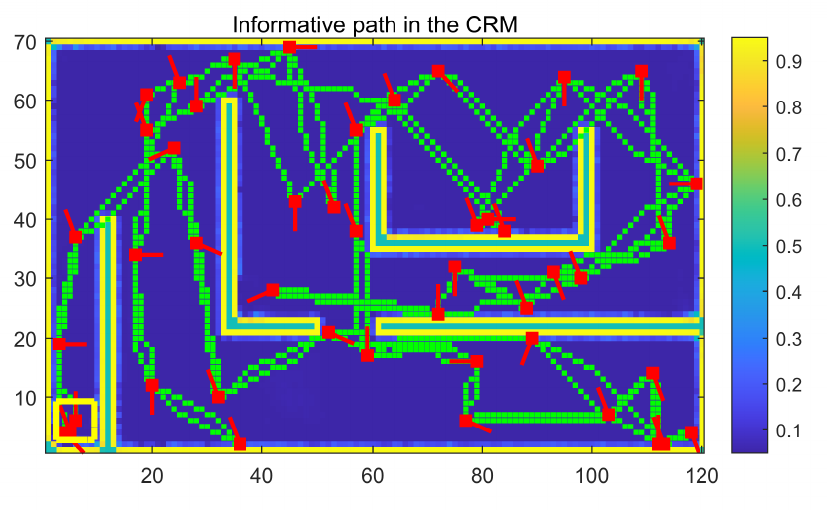}
			\end{minipage}%
		}%
		\subfigure[Unstructured synthetic map.]{
			\begin{minipage}[t]{0.5\linewidth}
				\centering
				\includegraphics[width=1.0\linewidth]{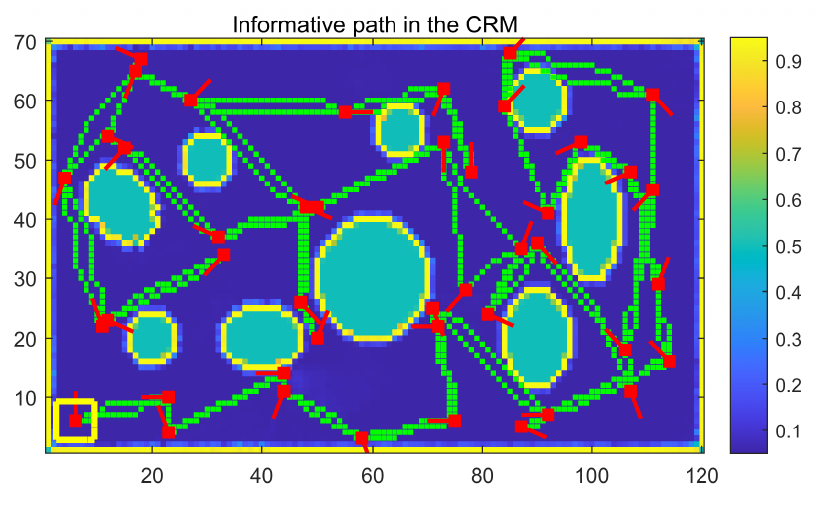}
			\end{minipage}%
		}%
		\centering
		\caption{Examples of CRMI-based robot exploration in unknown synthetic environments.  \textit{Yellow square}: the start and end points. The red line starting with a red square represents a planned robot action/pose in SE(2), and \textit{green square} is the local path between two actions.}
		\label{fig:maze-traj}
	\end{figure}
	\subsection{Synthetic Environments and Dataset Results}
	\label{ssec:simu}
	We run 20 Monte Carlo {(MC)} trials for all numerical simulations and dataset experiments on a laptop PC with a 2.6 GHz Intel i5-4210M CPU and 8G RAM. The robot poses are obtained by Gmapping \cite{grisetti2007improved}. 
	{Note that the man-made binary ground truth maps are used only to simulate the true world and generate virtual sensor observations for robot localization and mapping, and the robot cannot access these maps for exploration purposes, i.e., it still faces \textit{unknown environments}.}

	The minimum information threshold is $I_{\mathrm{th}} = 2$ bits, and the trade-off factor is $\alpha=1$. The $\lambda_z$ and $\lambda_m$ for CRMI calculation are both 0.1. 
	We also choose the parameters of $\zeta=0.001$ and $\sigma_{\theta}=100$ for the BKIO method.  
	We use 30 points (each point with eight directions uniformly) to generate candidate actions for each method, i.e., $N+N_q=240, N_{q}=2N$. 
	{We also set $N_{\mathrm{epoch}}=30$ for GPBO and BKIO methods. This value is chosen particularly and validated in an experimental study. More related details will be particularly analyzed below.}
	Note that we set an extreme case of numerous candidate actions to test the time-saving performance of the proposed methods.
	
\begin{figure}[ht]
		\centering
		\subfigure[Map entropy (structured)]{
			\begin{minipage}[t]{0.5\linewidth}
				\centering
				\includegraphics[width=1.8in]{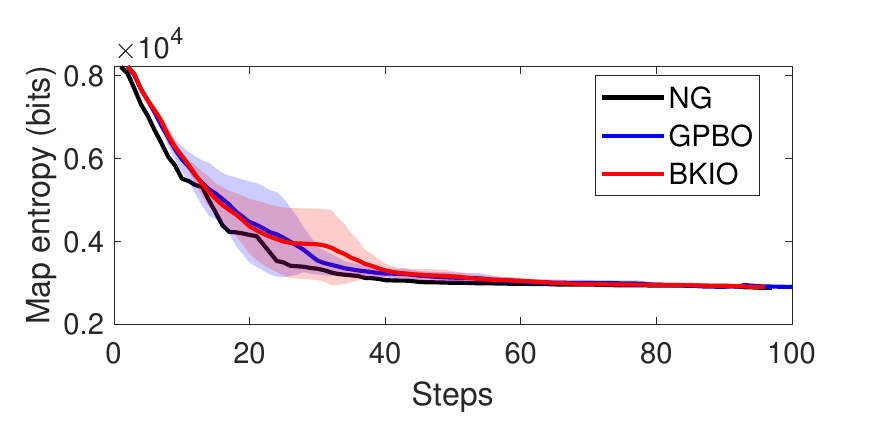}
			\end{minipage}%
		}%
		\subfigure[Coverage rate (structured)]{
			\begin{minipage}[t]{0.5\linewidth}
				\centering
				\includegraphics[width=1.8in]{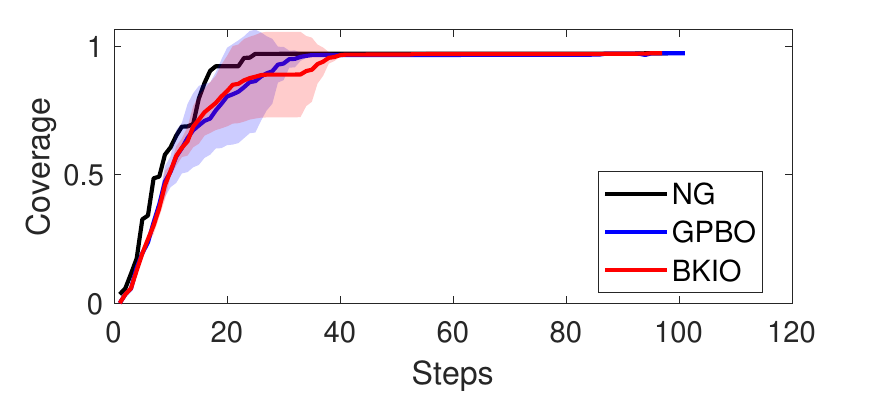}
			\end{minipage}%
		}%
		\quad
		\subfigure[Map entropy (unstructured)]{
			\begin{minipage}[t]{0.5\linewidth}
				\centering
				\includegraphics[width=1.8in]{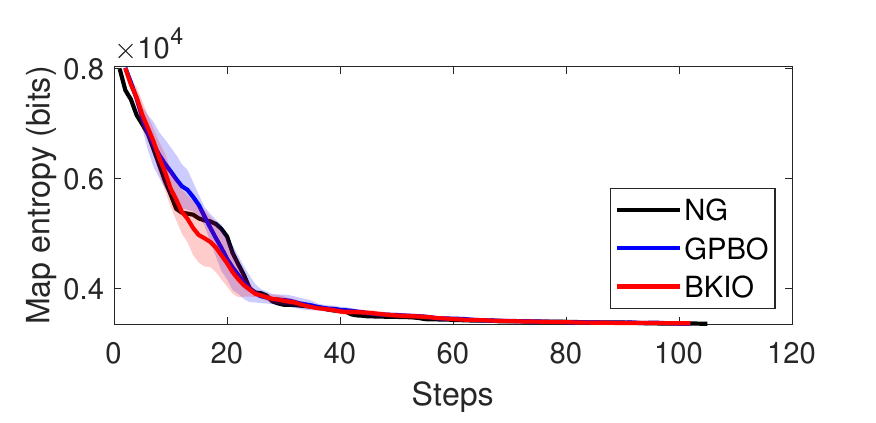}
			\end{minipage}%
		}%
		\subfigure[Coverage rate (unstructured)]{
			\begin{minipage}[t]{0.5\linewidth}
				\centering
				\includegraphics[width=1.8in]{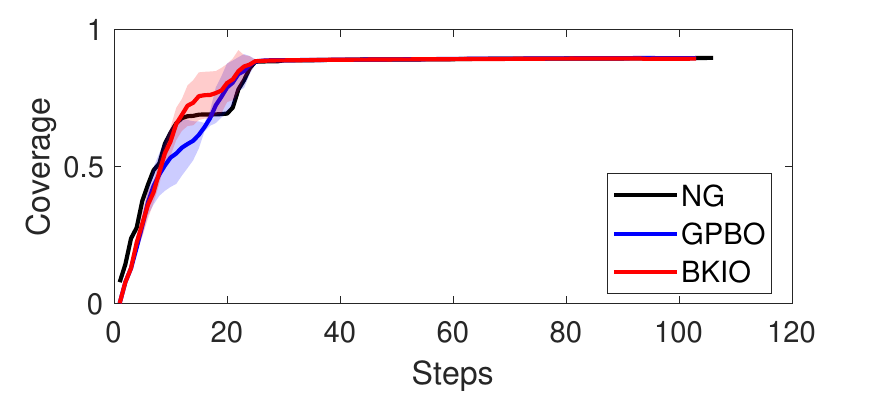}
			\end{minipage}%
		}%
		\centering
		\caption{Map entropy and coverage results of synthetic structured and unstructured maps. BKIO method can achieve similar exploration performance as GPBO, even better than GPBO in the unstructured map, and with less total exploration time.}
		\label{fig:maze-res}
	\end{figure}

	{To compare the exploration performance using different methods more quantitatively, referring to \cite{charrow2015information,zhang2020fsmi}, we prefer to use map entropy and coverage rate of MC results since there exists randomness in the exploration process.} 
	In the map entropy and coverage rate figures {(c.f. Fig.~\ref{fig:maze-res} and Fig.~\ref{fig:sea-res})}, the solid lines depict the means of {MC} trials {of each method, and the shaded regions represent the standard deviations. Here we omit the standard deviation of NG for the figure's readability.} 
	
	\textbf{Results from structured/unstructured cases}: 
	To simulate indoor and field scenes, we generate two synthetic maps with the size of $24~\mathrm{m} \times 14 ~\mathrm{m}$, one structured maze map with several walls (Fig.~\ref{fig:maze-traj}(a), $N_{\mathrm{loop}}$ = 100), and one unstructured map consisting of circles and ellipses {representing forest-like scene} (Fig.~\ref{fig:maze-traj}(b), $N_{\mathrm{loop}}$ = 150). {The unstructured one will obstruct the robot' FOV and need more exploration steps.}
	The map resolutions are both 0.2 m. The simulated range sensor has a FOV of $\pm$1.5 rad with a resolution of 0.05 rad and a maximum sensing range of $6$~m. The robot is initially at [1.2, 1.2]~m with 0~rad heading and trying to explore the prior unknown map. 
	
	The examples of resulting informative paths maximizing CRMI values are shown in Fig.~\ref{fig:maze-traj}.
	The quantitative results of structured and unstructured maps are shown in Fig.~\ref{fig:maze-res}. 
	BKIO and GPBO have similar evolution trends of map entropy reduction and coverage rate as NG in structured and unstructured maps, indicating the relatively low exploration performance loss in these BO-based methods.
	
\begin{figure}
		\centering
		\subfigure[Informative path (Seattle map)]{
			\begin{minipage}[t]{0.5\linewidth}
				\centering
				\includegraphics[width=0.8\linewidth]{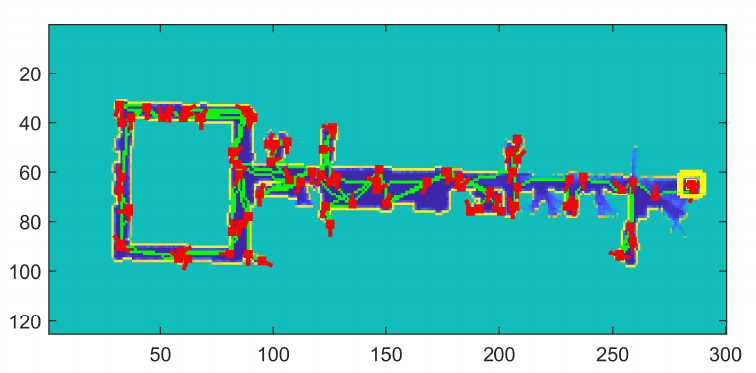}
			\end{minipage}%
		}%
		\subfigure[Total time cost (Seattle map)]{
			\begin{minipage}[t]{0.5\linewidth}
				\centering
				\includegraphics[width=1\linewidth]{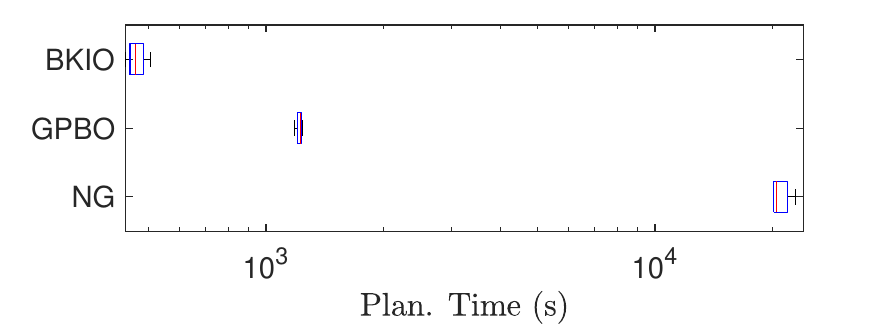}
			\end{minipage}%
		}%
		\centering
		\caption{BKIO-based robot exploration in the Seattle map. \textit{Yellow square}: start and end points. {Note that the $y$-axis uses log10 representation.}}
		\label{fig:real-exp}
	\end{figure}

\begin{figure}[ht]
		\centering
		\subfigure[Map entropy rate (Seattle map)]{
			\begin{minipage}[t]{0.45\linewidth}
				\centering
				\includegraphics[width=1.8in]{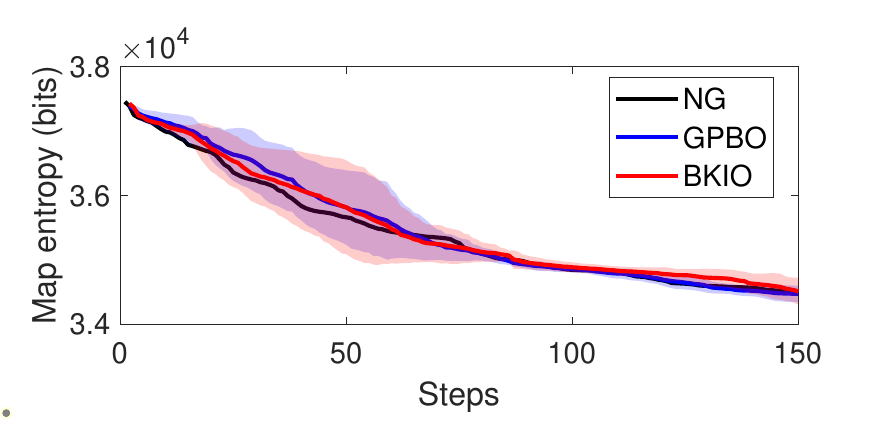}
			\end{minipage}%
		}%
		\subfigure[Coverage rate (Seattle map)]{
			\begin{minipage}[t]{0.45\linewidth}
				\centering
				\includegraphics[width=1.8in]{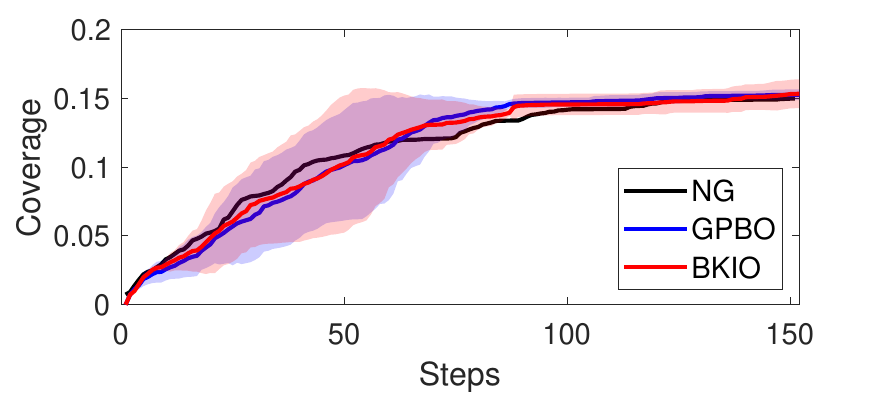}
			\end{minipage}%
		}%
		\centering
		\caption{Map entropy and coverage results of the Seattle map. Note that the entropy and coverage would not be 0 bit and 100\%, respectively, since plenty of thick walls and obstacles are also counted, not only the inside free areas, as in Fig.~\ref{fig:real-exp}(a).}
		\label{fig:sea-res}
	\end{figure}

	\textbf{Results from dataset case}: 
	To test our methods in a more realistic environment, we use a $24~\mathrm{m} \times 14 ~\mathrm{m}$ Seattle map \cite{Radish} containing narrow long corridors and cluttered rooms in Fig.~\ref{fig:real-exp}(a). To implement the exploration methods, we use the same simulated laser scanner as in the synthetic maps. 
	The initial robot pose is $x_0=[11.4, 2.6, -\pi/2]$. 
	Fig.~\ref{fig:real-exp} (a) selects one resulting informative path in Monte Carlo experiments.
	In the Seattle map, Fig.~\ref{fig:sea-res} shows BKIO and GPBO methods have similar reduction rates of map entropy and coverage rates w.r.t exploration steps. 
	The simulation results provide evidence BKIO and GPBO can achieve comparable exploration performance in different environments.
	
	\begin{figure}
		\centering
		\subfigure[Total time cost (structured)]{
			\begin{minipage}[t]{0.5\linewidth}
				\centering
				\includegraphics[width=1\linewidth]{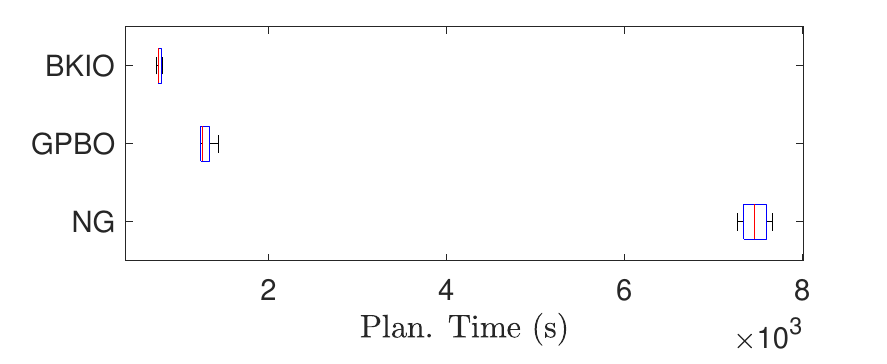}
			\end{minipage}%
		}%
		\subfigure[Total time cost (unstructured)]{
			\begin{minipage}[t]{0.5\linewidth}
				\centering
				\includegraphics[width=1\linewidth]{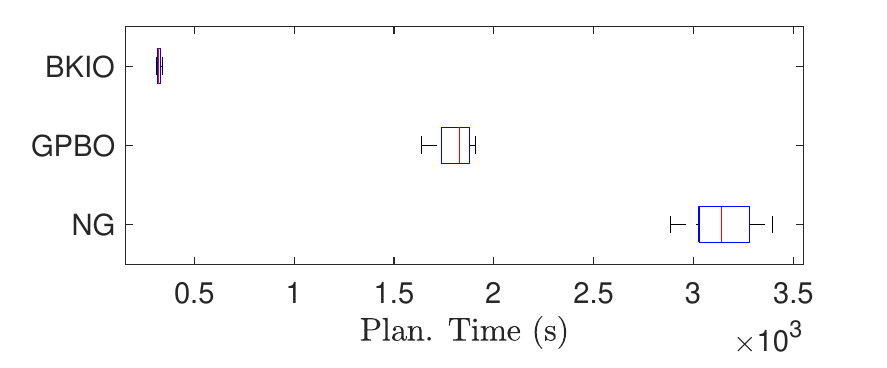}
			\end{minipage}%
		}%
		\centering
		\caption{{Exploration time cost using different methods in synthetic maps.} }
		\label{fig:simutime}
	\end{figure}
	
	\textbf{Time efficiency analysis:} 
	As in Table~\ref{tab:time}, we further analyze the time costs of both CRMI evaluation and exploration per iteration in each method.
	Both BKIO and GPBO are much more efficient than the NG method in each step, and both the time costs of CRMI evaluation and exploration get a significant reduction. 
	In addition, BKIO shows higher time efficiency performance than GPBO in all scenes.
	
	For example, compared to the NG method, BKIO decreases the information gain evaluation time by about 90.4\% in the structured map, compared with the one of 74.4\% using GPBO. 
	These performance differences still prevail in unstructured and large cluttered maps, e.g., BKIO decreases the evaluation time costs per step by about 92.6\% and 93.9\% in the more complex maps of unstructured and Seattle, much better than 76.1\% and 77.4\% of GPBO, respectively.
	The box plots depicting the entire exploration time cost in Fig.~\ref{fig:simutime} and Fig.~\ref{fig:real-exp}(b) also support the above efficiency analysis.
	\begin{table*}
		\caption{Time cost comparison per step of different methods}
		\begin{center}
			\begin{tabular}{c c c c}
				\hline
				Methods & Synthetic structured map & Synthetic unstructured map & Seattle map \cite{Radish} \\ \hline
				NG \cite{yang2021crmi}  & 28.31$\pm$1.49~/~51.05$\pm$2.66 & 40.59$\pm$3.04~/~71.72$\pm$3.27  & 53.02$\pm$5.52~/~101.15$\pm$3.11 \\ 
				GPBO & 7.25$\pm$0.98~/~10.27$\pm$1.07 & ~9.70$\pm$1.31/~17.08$\pm$2.24 & 11.97$\pm$1.10~/~21.65$\pm$1.94   \\ 	 
				\textbf{BKIO} & \textbf{2.71$\pm$0.36~/~5.14$\pm$0.48 } & \textbf{~2.98$\pm$0.43/~9.95$\pm$1.17 } & \textbf{3.22$\pm$0.42~/~10.54$\pm$1.09 } \\ 
				\hline
				\multicolumn{4}{l}{Note: Time cost (in seconds) of CRMI evaluation per step / Time cost of exploration per step.}
			\end{tabular}
			\label{tab:time}
		\end{center}
		\vspace{-0.5cm}
	\end{table*}
	
	{
		\textbf{Experimental study of different $N_{\mathrm{epoch}}$ values:}
		Here we will investigate the influence of $N_{\mathrm{epoch}}$ on time efficiency and quality, presenting the experimental study results in synthetic and dataset simulations, including some results about time efficiency using different $N_{\mathrm{epoch}}$.}
	
	{To study this, we use three representative $N_{\mathrm{epoch}}$ values of 1, 30, and 60 for BKIO, where $N_{\mathrm{epoch}}=1$ means the BO is only conducted once (i.e., only use BKI itself, denoted as `BKIO-1'), $N_{\mathrm{epoch}}=30$ is what we use in this paper (`BKIO-30'), and $N_{\mathrm{epoch}}=60$ means the BO is conducted excessively (`BKIO-60').}
	
	{We present the results of time efficiency in the following Table \ref{tab:bkio} and the exploration performance (map entropy reduction and coverage rate) in the following figures (Fig. \ref{fig:r1-ab}). The results show that though the time efficiency of BKIO-1 significantly outperforms BKIO-30 and BKIO-60, the resulting exploration performance of BKIO-1 is the worst of all in the following Fig. \ref{fig:r1-ab}, ending the exploration too early. }
	
	{BKIO-60 shows the best exploration performance but spends much more time than BKIO-30, while BKIO-30 performs similarly to BKIO-60, achieving a better balance of time efficiency and exploration performance. This phenomenon also implies that the performance improvement of higher $N_{\mathrm{epoch}}$ value has a marginal effect. }
	
	{To balance the time efficiency and exploration performance, we choose the  $N_{\mathrm{epoch}}=30$ for BKIO in the following simulations and real-world experiments.}
	
\begin{figure}[ht]
		\centering
		\subfigure[Map entropy (structured)]{
			\begin{minipage}[t]{0.5\linewidth}
				\centering
				\includegraphics[width=1\linewidth]{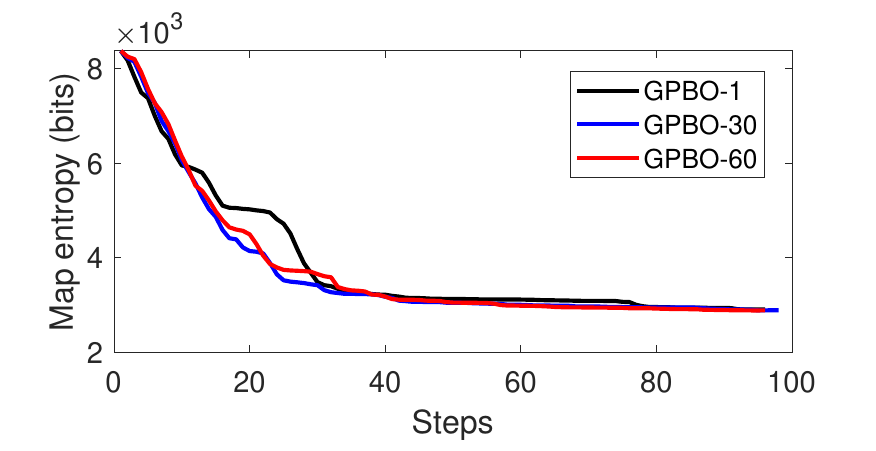}
			\end{minipage}%
		}%
		\subfigure[Coverage rate (structured)]{
			\begin{minipage}[t]{0.5\linewidth}
				\centering
				\includegraphics[width=1\linewidth]{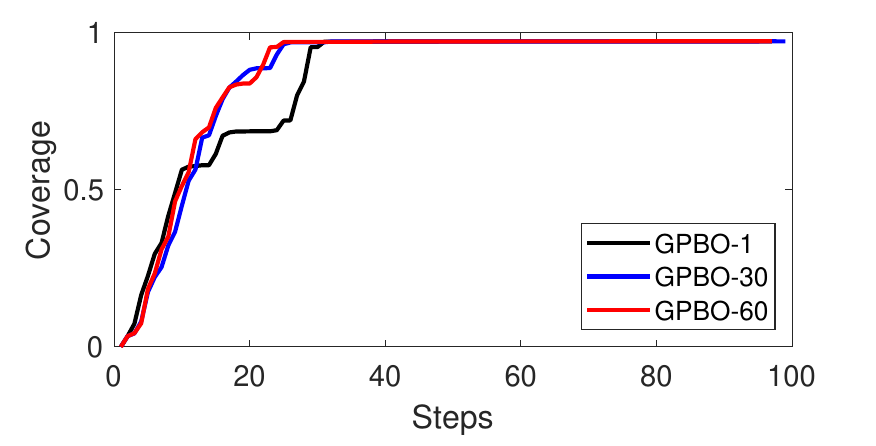}
			\end{minipage}%
		}%
		
		\subfigure[Map entropy (unstructured)]{
			\begin{minipage}[t]{0.5\linewidth}
				\centering
				\includegraphics[width=1\linewidth]{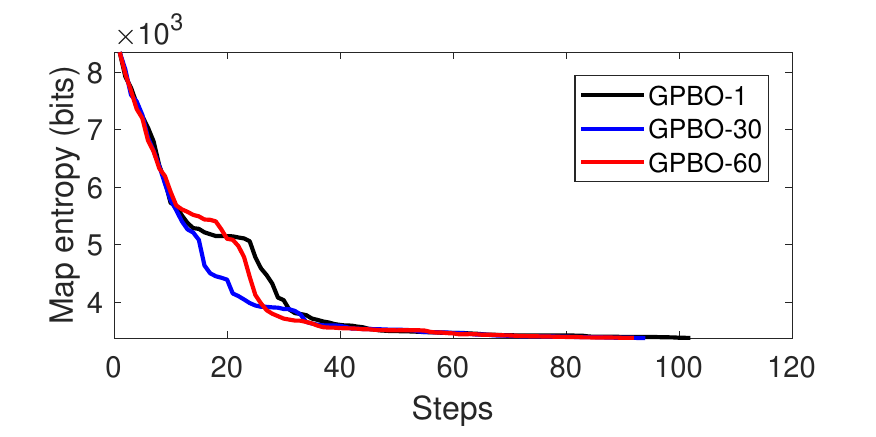}
			\end{minipage}%
		}%
		\subfigure[Coverage rate (unstructured)]{
			\begin{minipage}[t]{0.5\linewidth}
				\centering
				\includegraphics[width=1\linewidth]{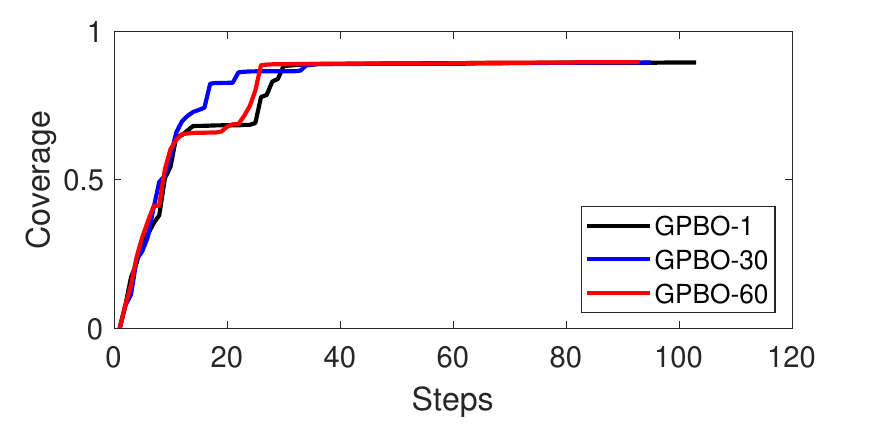}
			\end{minipage}%
		}%
		\centering
		\caption{GPBO exploration performance using different $N_{\mathrm{epoch}}$ values.}
		\label{fig:r1-ab-gp}
	\end{figure}

	{We have also studied how $N_{\mathrm{epoch}}$ influences GPBO using the same value settings. Similarly, the results in Fig. \ref{fig:r1-ab-gp} and Table \ref{tab:gpbo} still show that $N_{\mathrm{epoch}}=30$ is a better choice in practice.}
	
	\begin{figure}[ht]
		\centering
		\subfigure[Map entropy (structured)]{
			\begin{minipage}[t]{0.5\linewidth}
				\centering
				\includegraphics[width=1\linewidth]{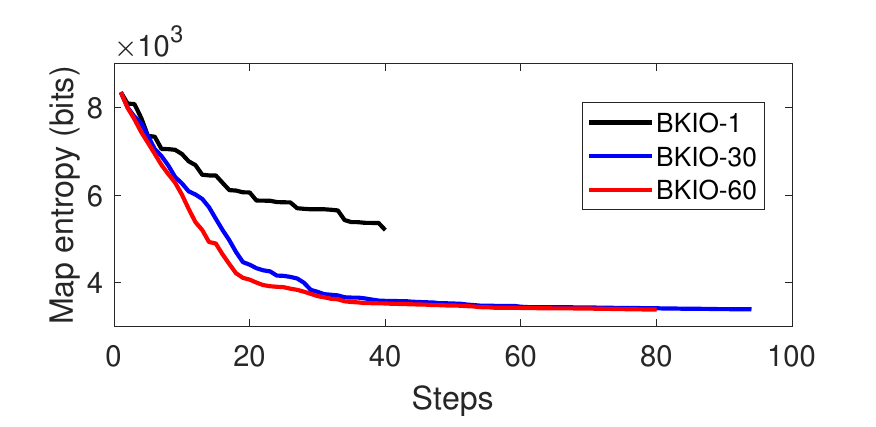}
			\end{minipage}%
		}%
		\subfigure[Coverage rate (structured)]{
			\begin{minipage}[t]{0.5\linewidth}
				\centering
				\includegraphics[width=1\linewidth]{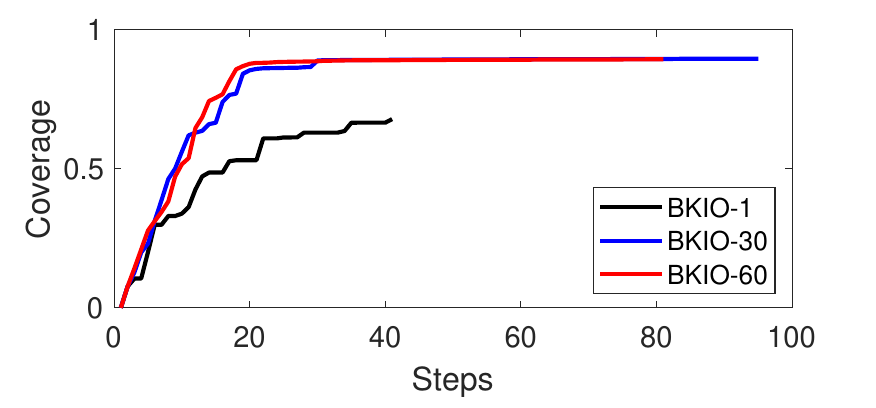}
			\end{minipage}%
		}%
		
		\subfigure[Map entropy (unstructured)]{
			\begin{minipage}[t]{0.5\linewidth}
				\centering
				\includegraphics[width=1\linewidth]{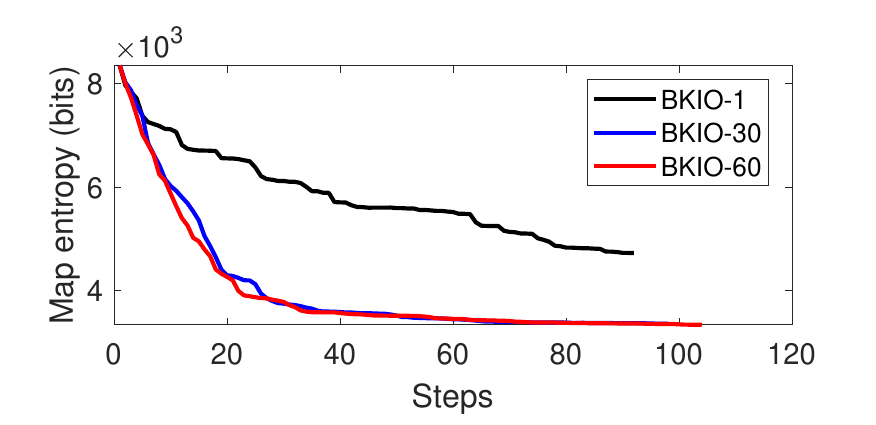}
			\end{minipage}%
		}%
		\subfigure[Coverage rate (unstructured)]{
			\begin{minipage}[t]{0.5\linewidth}
				\centering
				\includegraphics[width=1\linewidth]{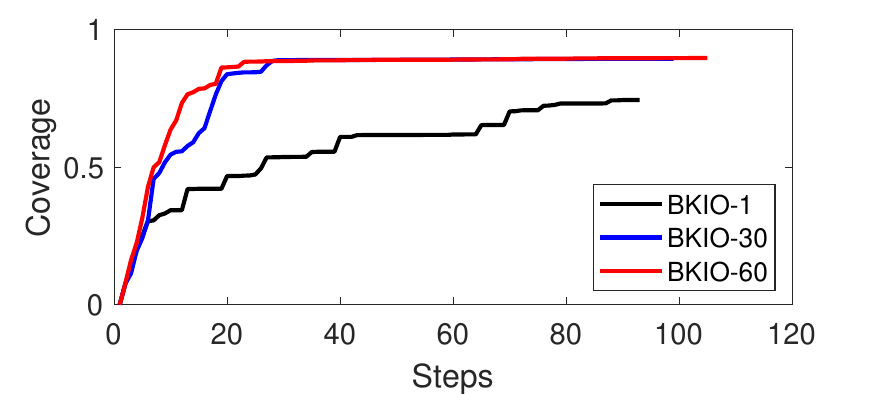}
			\end{minipage}%
		}%
		\centering
		\caption{BKIO exploration performance using different $N_{\mathrm{epoch}}$ values.}
		\label{fig:r1-ab}
	\end{figure}
	
	\begin{table}[ht]
		\caption{BKIO time cost (in sec) using different $N_{\mathrm{epoch}}$ values.}
		\begin{center}
			\begin{tabular}{c c c}
				\hline
				Methods & Structured map & Unstructured map \\ \hline
				BKIO-1  & 0.08~/~2.47 & 0.14~/~2.75  \\ 
				\textbf{BKIO-30} & \textbf{2.27~/~5.05} & \textbf{~3.07~/~9.61} \\ 	 
				BKIO-60 & {4.62~/~8.85 } & {7.96/~12.19 } \\ 
				\hline
				\multicolumn{3}{l}{Note: CRMI evaluation/exploration costs per step.}
			\end{tabular}
			\label{tab:bkio}
		\end{center}
		\vspace{-0.5cm}
	\end{table}
	
	\begin{table}[ht]
		\caption{GPBO time cost (in sec) using different $N_{\mathrm{epoch}}$ values.}
		\begin{center}
			\begin{tabular}{c c c}
				\hline
				Methods & Structured map & Unstructured map \\ \hline
				GPBO-1  & 0.23~/~3.84 & 0.44~/~4.82  \\ 
				\textbf{GPBO-30} & \textbf{6.57~/~11.36} & \textbf{~7.65~/~15.96} \\ 
				GPBO-60 & {9.36~/~16.33 } & {9.31/~20.04 } \\ 
				\hline
				\multicolumn{3}{l}{Note: CRMI evaluation/exploration costs per step.}
			\end{tabular}
			\label{tab:gpbo}
		\end{center}
		\vspace{-0.5cm}
	\end{table}

	{
		\textbf{Local trade-off analysis:} 
		Here we use an intuitive example in the structured scene to show how the UCB of Eq.~\eqref{eq:ucb} improves the trade-off between exploration and exploitation locally at each exploration step. }
	
\begin{figure}
		\includegraphics[width=1.0\linewidth]{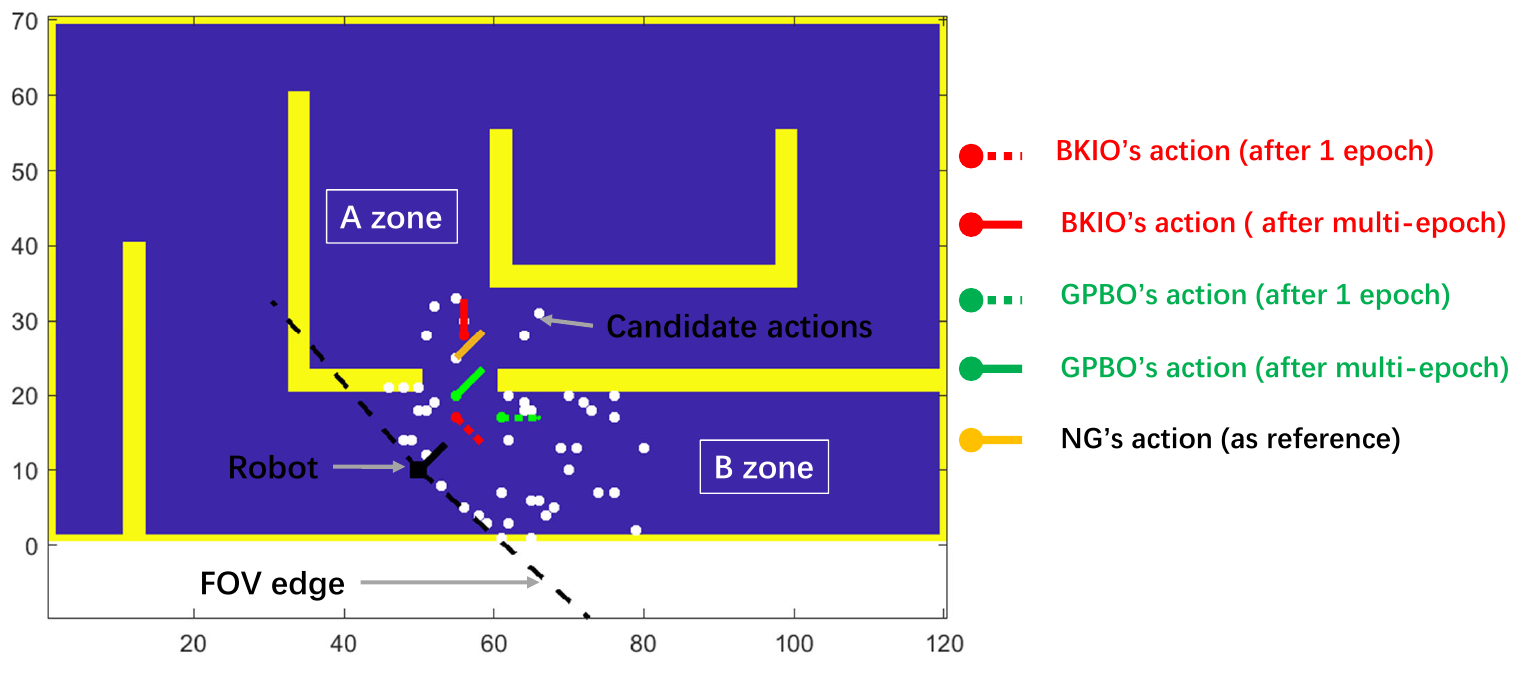}
		\centering
		\caption{{Illustration of the trade-off between exploration and exploitation.}}
		\label{fig:trade}
	\end{figure}

	{As in Fig.~\ref{fig:trade}, the robot stands (black square with a black bar) at a fork road and decides where to go next. The NG method evaluates all candidate actions and suggests the best action (orange bar, max CRMI: 76.95 bits) moving towards the A zone. In contrast, the Bayesian optimization-based GPBO and BKIO methods suggest different actions before and after multiple epoch optimization. GPBO and BKIO initially suggested the action towards the B zone, marked by the green and red dashed lines, respectively (CRMI values: $65.12$ and $67.57$). }
	{After multiple epoch optimization, the results of GPBO and BKIO (green and red bars, respectively) have been improved (CRMI values: $71.69$ and $73.44$) and got closer to NG's suggested action. 
		This means that, given limited explicitly evaluated samples, BO-based methods can avoid being stuck in the local optimality of max CRMI and explores actions with high prediction uncertainty, i.e., the multi-epoch BO chooses the promising actions suggested by the UCB function (Eq.~\eqref{eq:ucb}) and evaluates them explicitly, then adds them to the training set and improves the surrogate predictive model between actions and their CRMI. }
	
	{Finally, using several epochs, the predicted results get closer to the ground truth, i.e., the local trade-off between exploration and exploitation in CRMI inference at each step can be achieved using the UCB function.}
	
	\subsection{Real-world Experiments}
	\label{ssec:real}
	We deploy our proposed BKIO and GPBO on the Fetch \footnote{http://docs.fetchrobotics.com} mobile robot running C++/Python code in the Robot Operating System (ROS). As shown in Fig.~\ref{fig:fetch}, we conduct experiments in an approximately $25~\mathrm{m} \times 35 ~\mathrm{m}$ real-world office environment in Level 11, Building 2 of the University of Technology Sydney. There are many chairs, tables, and glass walls in the office and kitchen zones, which would hinder the ego perception and motion of the robot. The FOV of the LiDAR is 220 degrees, and the max sensing range is 25 m. Note that the robot explores the environment autonomously and only uses the laser scan data from the SICK 2D LiDAR (15 Hz) without cameras. The max sensing range is set to 8 m, forcing this `sensing-limited' robot to explore longer. 
	
	Here we use the Cartographer \cite{hess2016real} as the localization module and CRM \cite{agha2019confidence} as the mapping module in the exploration framework. The robot calls for a self-defined ROS service to evaluate the CRMI of each action generated at the extracted frontiers, then decides where to go at the next step until the exploration ends {when the travel cost reaches 500 m or the expected CRMI of the best action is lower than the predefined threshold of 10 bits.}
	An example video is available here\footnote{https://youtu.be/sOW4fuaAwT8}.
	
\begin{figure}[ht]
		\centering
		\subfigure[Map entropy (w.r.t distance)]{
			\begin{minipage}[t]{0.5\linewidth}
				\centering
				\includegraphics[width=1.8in]{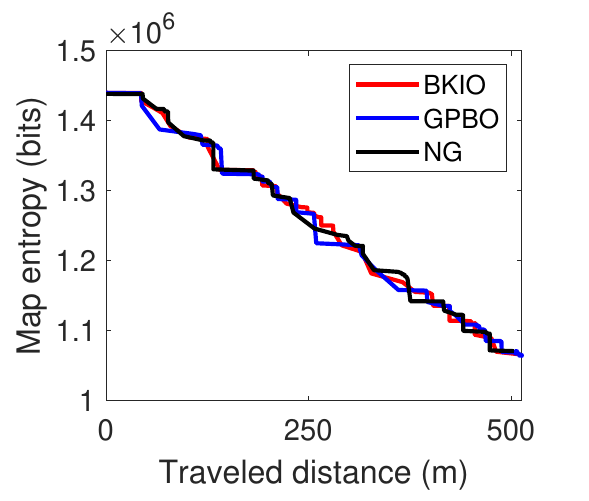}
			\end{minipage}%
		}%
		\subfigure[Covered area (w.r.t distance)]{
			\begin{minipage}[t]{0.5\linewidth}
				\centering
				\includegraphics[width=1.8in]{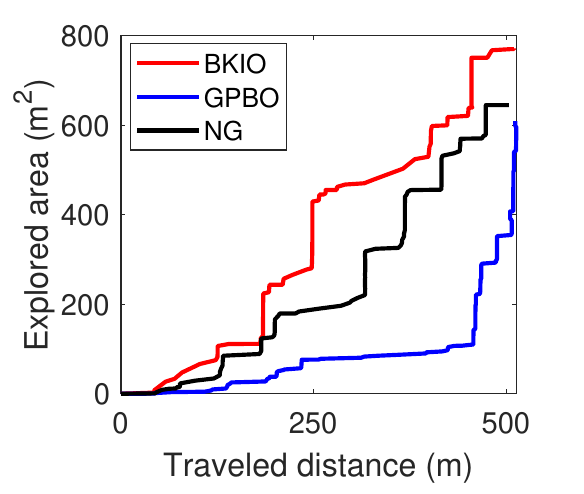}
			\end{minipage}%
		}%
		\quad
		\subfigure[Map entropy (w.r.t time)]{
			\begin{minipage}[t]{0.5\linewidth}
				\centering
				\includegraphics[width=1.8in]{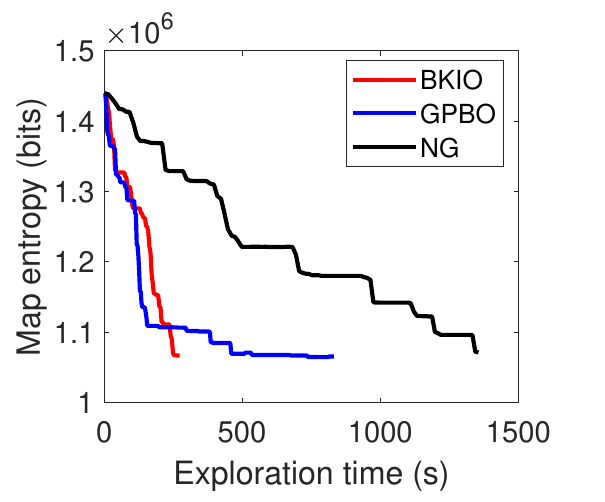}
				\label{fig:uts-res-c}
			\end{minipage}%
		}%
		\subfigure[Covered area (w.r.t time)]{
			\begin{minipage}[t]{0.5\linewidth}
				\centering
				\includegraphics[width=1.8in]{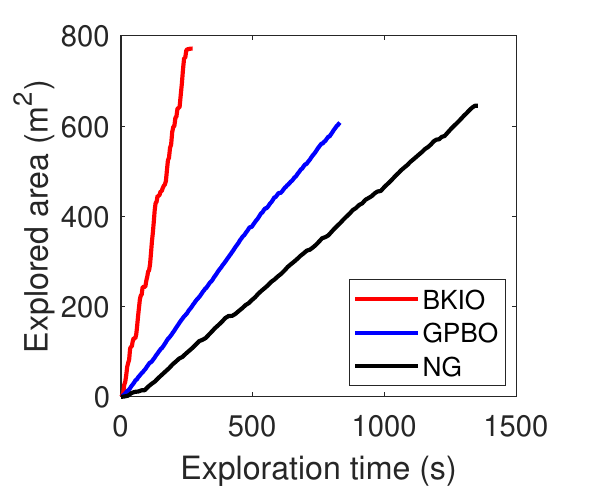}
				\label{fig:uts-res-d}
			\end{minipage}%
		}%
		\centering
		\caption{{Real-world map entropy and coverage results w.r.t robot travel distance and total time cost, respectively.}}
		\label{fig:uts-res}
	\end{figure}

	Fig.~\ref{fig:uts-res} shows the comparative evolution results of map entropy and explored areas using three different methods {as the travel distance and total exploration time increase.} 
	The map entropy and coverage curves in Fig.~\ref{fig:uts-res}(a) and (b) show these methods all complete the exploration finally.
	{In Fig~\ref{fig:uts-res}(a), the map entropy curves of the three methods w.r.t the traveled distance are similar, but the explored areas vary a lot in Fig.~\ref{fig:uts-res}(b). This is mainly because we determine a map cell as already explored if its expected occupancy deviates from 0.5, otherwise unexplored (bool 0 or 1) when computing the covered area; but when computing the map entropy, we use the continuous expected occupancy (in the interval [0,1]). The rapid increase of explored areas in Fig.~\ref{fig:uts-res}(b) and Fig.~\ref{fig:uts-res-d} also implies our BKIO can drive the robot to explore more unknown areas rapidly, given a limited distance budget, and get an environmental sketch for the subsequent robot application.}
	
	{Fig.~\ref{fig:uts-res}(b) shows that the GPBO travels a longer distance at the start stage but gets the less explored area than others since it moves in the corridors for a long distance and explores more aggressively and roughly in the later stage.}
	{In Fig.~\ref{fig:uts-res-d}, we} can observe that the BKIO and GPBO methods outperform the NG in terms of exploration time cost significantly (79.9\% and 38.6\% reduction, respectively), where BKIO performs better in explored areas than the others. 
	Fig.~\ref{fig:uts-res}(c) and (d) also show our methods conduct much more efficient exploration in map uncertainty reduction and exploring unknown areas.
	
	Moreover, {as in Fig.~\ref{fig:uts-res-d},} BKIO reduces about 67.3\% total time cost to end the exploration and increases about 21\% covered areas than GPBO. 
	Further, the mean time costs of each exploration step are 12.2095$\pm$0.6383s, 0.2007$\pm$0.1362s, and 0.0038$\pm$0.0033s for NG, GPBO, and BKIO, respectively. This also evidences the advantage of our proposed methods.

\begin{figure}[ht]
		\includegraphics[width=1.0\linewidth]{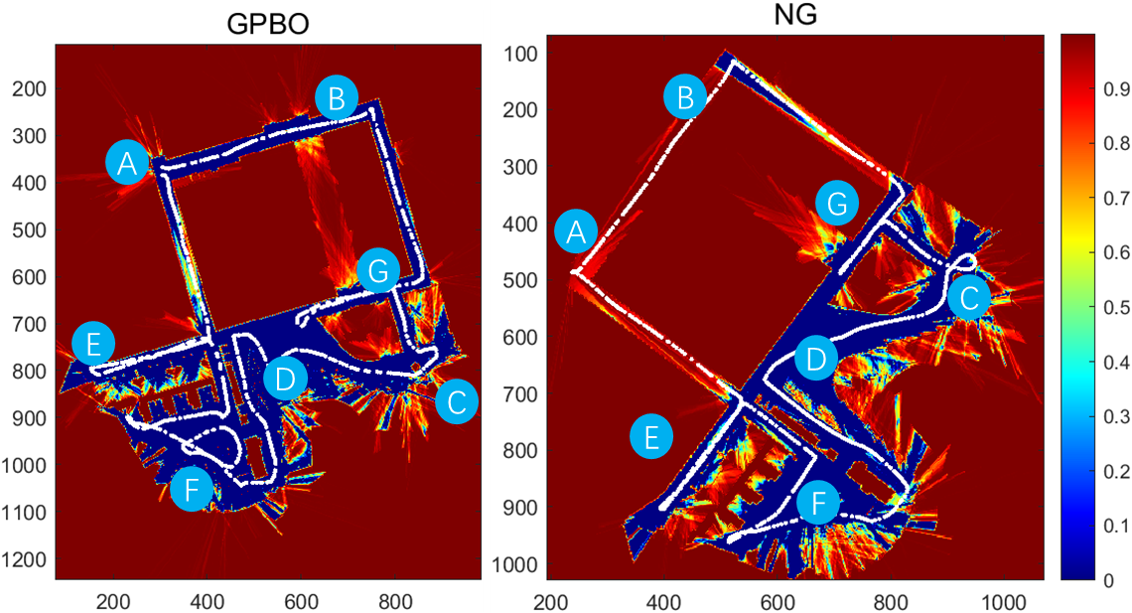}
		\centering
		\caption{GPBO and NG exploration results.}
		\label{fig:uts-gpbo}
	\end{figure}

	{The exploration sequence differences of the three methods are shown in Figure \ref{fig:uts-gpbo}, where the overall sequences of NG and GPBO in this figure are A-B-G-C-D-F-E-A and A-B-A-E-F-D-C-G-B-A, respectively. Both BKIO and NG began with A-B-G-C, while GPBO began with A-B-A-E.
		There are mainly two reasons accounting for this phenomenon. Firstly, the best actions suggested by different methods at each exploration step vary because the information evaluation accuracy differs, such as GPBO vs. BKIO (w.r.t NG as the ground truth). This difference accumulates over the exploration steps (e.g., choosing different intersections) and affects the resulting exploration sequence. 
		These different choices made by the three methods will significantly change the exploration sequence. This phenomenon also exists in the simulation; referring to the local trade-off analysis in Section IV-A, Fig. \ref{fig:trade} shows that all methods choose different actions at the intersection since the predicted CRMI values are unequal.}
	
	{Though, the overall exploration metrics such as map entropy, explored area, total time cost, travel distance, and average exploration time cost can still help us assess the performance of different methods.}
	{Secondly, for each method, the CRMI-based objective function values of several different actions will be equal in rare cases, such as multiple actions all facing the open unknown space or parallel to the wall. The best action can only be selected from these actions randomly; we also try to improve this in our next work, such as using adaptive sampling or improving frontier extraction.}
	
	\subsection{Discussion}
	As presented in previous experiments of typical scenes, our proposed methods (GPBO and BKIO) have shown desired exploration performance in efficiency and accuracy, especially the more efficient BKIO. 
	Moreover, we use the public packages (e.g., scikit-learn and NumPy) to implement GP, but the BKI code is non-optimized. Hence we believe our methods' above performance improvement trend is not biased and would be more salient when facing more complex scenes.
	
	The significant time-saving benefits for online exploration tasks greatly, especially when using computationally limited or energy-limited platforms. In other words, we offer two options for efficient information-based robot exploration tasks in unstructured and relatively large areas.
	
	Another thing we should clarify is that we are not aiming to propose a new exploration framework. The exploration framework we use in this paper (i.e. Algorithm~\ref{alg:bkiexp}) stems from depth-first search, it is just used to run BKIO, GPBO, and NG methods. In other words, we can use another exploration framework instead, such as {the sampling-based incrementally-informative graph} \cite{ghaffari2019sampling}. We have also tested our methods in both sampling-based (Section \ref{ssec:simu}) and frontier-based (Section \ref{ssec:real}) exploration. Hence we would not discuss the advantages and disadvantages of these frameworks because we focus on decreasing the evaluation time cost at each exploration step and the time cost of the whole exploration.
	
	Moreover, the information metric is also not limited to CRMI. One can apply our methods to other information metrics for robot exploration, such as CSQMI \cite{charrow2014approximate} and {GP-based MI (GPMI) \cite{ghaffari2019sampling}}. 
	
	{In some experiments (e.g.,  Fig.~\ref{fig:maze-res}), one can also find NG sometimes performs worse during the exploration (14-21 steps). The underlying reason is mainly that NG completely pursues the action maximizing the expected CRMI (i.e., \textit{exploitation}), but may neglect more promising actions that have less expected MI currently (i.e., \textit{exploration}). This myopic behavior may lead to undesired paths or even worse. Our future work will also study the non-myopic informative path planning in our next work to realize the global trade-off between exploration and exploitation.}

	\section{Conclusions}
	\label{sec:con}
	This paper mainly contributed to new learning-based approaches with high efficiency for information-theoretic robot exploration in unknown environments. In particular, an information gain evaluation method for inferring the CRMI of numerous sampled robot actions is proposed based on GP and BO. With the Bayesian kernel inference-based BO, the time complexity of CRMI prediction decreases to a logarithm level. An informative objective function integrating the predicted CRMI and uncertainty is also used to ensure the bounded regret and {local trade-off between exploration and exploitation at each step.}
	The proposed method also gets verified under an autonomous exploration framework by extensive simulations and real-world experiments in different scenes, which reveals the proposed BKIO and GPBO exploration methods outperform the non-learning ones overall in efficiency without losing much exploration performance, especially in unstructured and large cluttered scenes. This would also benefit other information-based exploration methods stuck by efficiency issues.
	Future work mainly involves the extension to 3D exploration.
	
	
	
	
	\bibliographystyle{IEEEtran}
	\bibliography{ral23_abbr.bib}
	
	\vfill
	
\end{document}